\documentclass[1p]{elsarticle}

\makeatletter
\def\ps@pprintTitle{%
    \let\@oddhead\@empty
    \let\@evenhead\@empty
    \let\@oddfoot\@empty
    \let\@evenfoot\@oddfoot
}
\makeatother

\usepackage{multirow}
\usepackage{enumitem}
\usepackage{todonotes}
\usepackage{xcolor}
\usepackage{amsmath,amsfonts,amssymb}
\usepackage{graphicx}
\usepackage{subcaption}
\usepackage{hyperref}

\hypersetup{
    colorlinks,
    linkcolor={red!50!black},
    citecolor={blue!50!black},
    urlcolor={blue!80!black}
}

\setlist{nolistsep}

\begin{document}

\begin{frontmatter}
\title{Extracting Contact and Motion from Manipulation Videos}

\address[mymainaddress]{University of Maryland at College Park, Maryland, 20742}
\address[cs_email]{\{kzampog,kganguly,yiannis\}@cs.umd.edu}
\address[umiacs_email]{fer@umiacs.umd.edu}
\cortext[mycorrespondingauthor]{Corresponding author}
\fntext[cs_affil]{University of Maryland Computer Science Department}
\fntext[umiacs_affil]{University of Maryland Institute for Advanced Computer Studies}

\author[mymainaddress,cs_email]{Konstantinos Zampogiannis\corref{mycorrespondingauthor}\fnref{cs_affil}}
\author[mymainaddress,cs_email]{Kanishka Ganguly\fnref{cs_affil}}
\author[mymainaddress,umiacs_email]{Cornelia Ferm{\"u}ller\fnref{umiacs_affil}}
\author[mymainaddress,cs_email]{Yiannis Aloimonos\fnref{cs_affil}}
\begin{abstract}
When we physically interact with our environment using our hands, we touch objects and force them to move: contact and motion are defining properties of manipulation.
In this paper, we present an active, bottom-up method for the detection of actor-object contacts and the extraction of moved objects and their motions in RGBD videos of manipulation actions.
At the core of our approach lies non-rigid registration: we continuously warp a point cloud model of the observed scene to the current video frame, generating a set of dense 3D point trajectories.
Under loose assumptions, we employ simple point cloud segmentation techniques to extract the actor and subsequently detect actor-environment contacts based on the estimated trajectories.
For each such interaction, using the detected contact as an attention mechanism, we obtain an initial motion segment for the manipulated object by clustering trajectories in the contact area vicinity and then we jointly refine the object segment and estimate its 6DOF pose in all observed frames.
Because of its generality and the fundamental, yet highly informative, nature of its outputs, our approach is applicable to a wide range of perception and planning tasks.
We qualitatively evaluate our method on a number of input sequences and present a comprehensive robot imitation learning example, in which we demonstrate the crucial role of our outputs in developing action representations/plans from observation.
\end{abstract}


\end{frontmatter}


\section{Introduction}\label{sec:intro}
A manipulation action, by its very definition, involves the handling of objects by an intelligent agent.
Every such interaction requires  physical contact between the actor and some object, followed by the exertion of forces on the manipulated object, which typically induce motion.
When we open a door, pick up a coffee mug, or pull a chair, we invariably touch an object  and cause it (or parts of it) to move.
This obvious observation demonstrates that \emph{contact} and \emph{motion} are two fundamental aspects of manipulation.

Contact and motion information alone are often sufficient to  describe  manipulations in a wide range of applications, as they  naturally encode crucial information regarding the performed action.
Contact encodes \emph{where} the affected object was touched/grasped, as well as \emph{when} and for how long the interaction took place.
Motion conveys  \emph{what} part of the environment (i.e. which object or object part) was manipulated and \emph{how} it moved.

The ability to automatically extract contact and object motion information from video either directly solves or can significantly facilitate a number of common perception tasks.
For example, in the context of manipulation actions, knowledge of the spatiotemporal extent of an actor-object contact automatically provides action \emph{detection/segmentation} in the time domain, as well as \emph{localization} of the detected action in the observed space \cite{poppe2010survey,weinland2011survey}.
At the same time, motion information bridges the gap between the observation of an action and its semantic grounding.
Knowing what part of the environment was moved effectively acts as an attention mechanism for the manipulated \emph{object recognition} \cite{rutishauser2004bottom,ba2014multiple}, while the extracted motion profile provides invaluable cues for \emph{action recognition}, in both ``traditional" \cite{wang2013action,poppe2010survey,weinland2011survey} and deep learning \cite{simonyan2014two} frameworks.

Robot imitation learning is   rapidly gaining attention.
The use of robots in less controlled workspaces and even domestic environments necessitates the development of easily applicable methods for robot ``programming": autonomous robots for  manipulation tasks must efficiently \emph{learn} how to manipulate.
Exploiting contact and motion information can largely automate robot replication of a wide class of actions.
As we will discuss later, the detected contact area can effectively bootstrap the grasping stage by guiding primitive fitting and grasp planning, while the extracted object and its motion capture the trajectory to be replicated as well as any applicable kinematic/collision constraints.
Thus, the components introduced in this work are  essential for  building complex, hierarchical models of action (e.g., behavior trees, activity graphs) as they appear in the recent literature \cite{kruger2011object,amaro2014understanding,summers2012using,yang2014cognitive,yang2015robot,aksoy2011learning,zampogiannis2015learning}.

In this paper, we present an unsupervised, bottom-up method for estimating from RGBD video the contacts and object motions in manipulation tasks.
Our approach is fully 3D and relies on dense motion estimation: we start by capturing a point cloud model of the observed scene and continuously warp/update it throughout the duration of the video.
Building upon our estimated dense 3D point trajectories, we use simple concepts and common sense rules to segment the actor and detect actor-environment contact locations and time intervals.
Subsequently, we exploit the detected contact to guide the motion segmentation of the manipulated object and, finally, estimate its 6DOF pose in all observed video frames.
Our intermediate and final results are summarized in Table \ref{tab:method_i_o}.

It is worth noting that we do not treat contact detection and object motion segmentation/estimation independently: we use the detected contact as an \emph{attention mechanism} to guide the extraction of the manipulated object and its motion.
This \emph{active} approach provides an elegant and effective solution to our motion segmentation task.
A passive approach to our problem would typically segment the whole observed scene into an \emph{unknown} (i.e. to be estimated) number of motion clusters.
By exploiting contact, we avoid having to solve a much larger and less constrained problem, while gaining significant improvements in terms of both computational efficiency and segmentation/estimation accuracy.

The generality of our framework, combined with the highly informative nature of our outputs, renders our approach applicable to a wide spectrum of perception and planning tasks.
In Section \ref{sec:approach}, we provide a detailed technical description of our method, while in Section \ref{sec:experiments}, we demonstrate our intermediate results and final outputs for a number of input sequences.
In Section \ref{sec:applications}, we present a comprehensive example of how our outputs were successfully used to facilitate a robot imitation learning task.

\section{Related Work}\label{sec:related}
We focus our literature review on recent works in four areas that are most relevant to the our twofold problem, and the major processes/components upon which we build.
We deliberately do not review works from the action recognition literature; while our approach may very appropriately become a component of a higher-level reasoning solution, the scope of this paper is the extraction of contacts, moving objects, and their motions.

\textbf{Scene flow.}
Scene flow refers to the dense 3D motion field of an observed scene with respect to a camera; its 2D projection onto the image plane of the camera is the optical flow.
Scene flow, analogously to optical flow, is typically computed from multi-view frame pairs \cite{yan2016scene}.
There have been a number of successful recent works on scene flow estimation from RGBD frame pairs, following both variational \cite{herbst2013rgb,quiroga2014dense,jaimez2015motion,jaimez2015primal,jaimez2017fast} and deep learning \cite{mayer2016large} frameworks.
While being of great relevance in a number of motion reasoning tasks, plain scene flow cannot be directly integrated into our pipeline, which requires \emph{model-to-frame} motion estimation: the scene flow motion field has a 2D support (i.e. the image plane), effectively warping the 2.5D geometry of an RGBD frame, while we need to appropriately warp a \emph{full} 3D point cloud model.

\textbf{Non-rigid registration.}
The non-rigid alignment of 3D point sets can be viewed as a generalization of scene flow, in the sense that the estimated motion field is supported by a 3D point cloud: the goal is to estimate point-wise transformations (usually rigid) that best align the point set to the target geometry under certain global prior constraints (e.g., `as-rigid-as-possible' \cite{sorkine2007rigid}).
The warp field estimation is performed either by iterating between correspondence estimation and motion optimization \cite{tam2013registration,amberg2007optimal,newcombe2015dynamicfusion,innmann2016volume}, or in a correspondence-free fashion, by aligning volumetric SDFs (Signed Distance Fields) \cite{slavcheva2017killingfusion}.
For this work, and due to lack of publicly available solutions, we have implemented a non-rigid registration algorithm similar to \cite{newcombe2015dynamicfusion} and \cite{innmann2016volume} (Section \ref{sec:non_rigid_icp}) and released it as part of our \texttt{cilantro} \cite{cilantro} library.

\textbf{Contact detection.}
A CNN-based method for grasp recognition is introduced  in \cite{yang2015grasp}.
A 2D approach for detecting ``touch" interactions between a caregiver and an infant is presented in \cite{chen2016touch}.
To the best of our knowledge, there is no prior work on explicitly determining the spatiotemporal extent of human-environment contact.

\textbf{Motion segmentation.}
A very large volume of works on motion segmentation have casted the problem as subspace clustering of 2D point trajectories, assuming an affine camera model \cite{yan2006general,tron2007benchmark,costeira1995multi,kanatani2001motion,rao2010motion,vidal2004motion}.
In \cite{katz2013interactive}, an active approach for the segmentation and kinematic modeling of articulated objects is proposed, which relies on  the robot manipulation capabilities to induce object motion.
In \cite{herbst2012object}, object segmentation is performed from two RGBD frames, one before and one after the manipulation of the object, by rigidly aligning and `differencing' the two views and robustly estimating rigid motion between the `difference' regions.
The same method is used in \cite{herbst2013rgb}, where scene flow is used to obtain motion proposals, followed by an MRF inference step.
In \cite{ruenz2017icra}, joint tracking and reconstruction of multiple rigidly moving objects is achieved by combining two segmentation/grouping strategies with multiple surfel fusion \cite{whelan2015elasticfusion} instances.
A naive integration of a generic motion segmentation algorithm for the extraction of the manipulated object into our pipeline would be suboptimal in multiple ways.
For instance, given the fact that there may exist an unknown number of other object motions that are irrelevant to the manipulation, we would be solving an unnecessarily hard problem.
For the same reason, we would have little control over the segmentation granularity, which could cause the manipulated object to be over/under-segmented.
Instead, we leverage the detected contact and bootstrap our segmentation by an informed trajectory clustering approach that is similar to \cite{ochs2014segmentation}.

\section{Our Approach}\label{sec:approach}
\subsection{Overview}
\begin{figure}[!ht]
	\centering
	\includegraphics[width=0.98\textwidth]{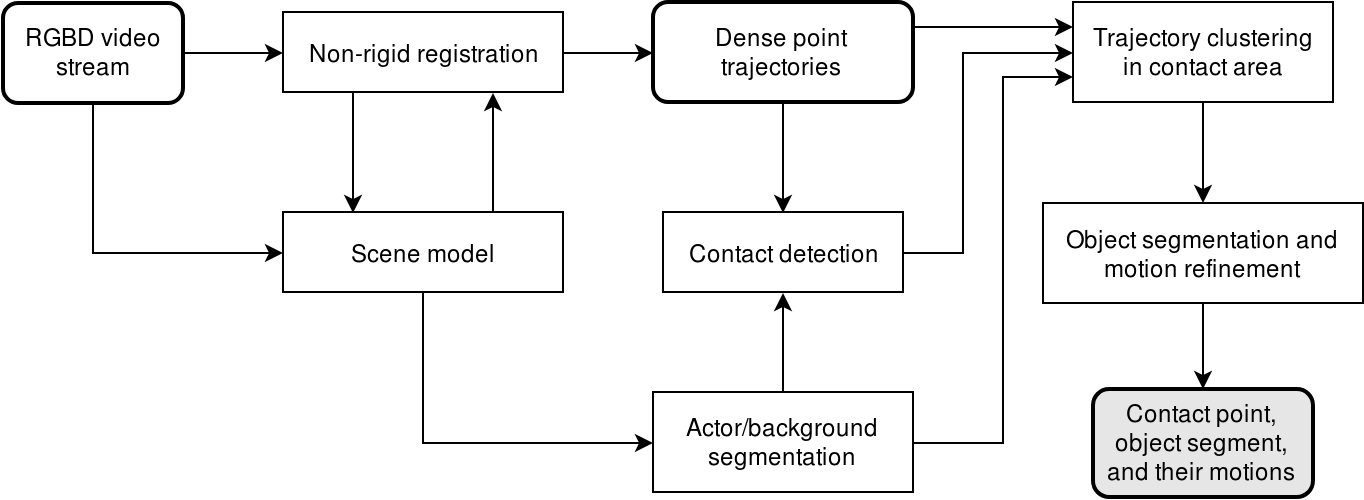}
    \caption{A high-level overview of our modules and their connections in the proposed pipeline.}
    \label{fig:overview}
\end{figure}

We present an automated system that, given a video of a human performing a manipulation task as input, \emph{detects} and \emph{tracks} the parts of the environment that participate in the manipulation. More specifically, our system is able to visually detect physical contact between the actor and their environment, and, using contact as an attention mechanism, eventually segment the manipulated object and estimate its 6DOF pose in every observed video frame.
Our pipeline, as well as the interactions of the involved processes, are sketched in Fig. \ref{fig:overview} and followed by a more detailed description. An in-depth discussion of our core modules is provided in the following subsections.

The input to our system is an RGBD frame sequence, captured by a commodity depth sensor, of a human actor performing a task that involves the manipulation of objects in their environment. We assume that the input depth images are registered to and in sync with their color counterparts. Using estimates of the color camera intrinsics (e.g., from the manufacturer provided specifications), all input  RGBD frames are back-projected to 3D point clouds (colored, with estimated surface normals), on which all subsequent processing is performed.

At the core of our method lies non-rigid point cloud registration, described in detail in Section \ref{sec:non_rigid_icp}. An initial point cloud model of the observed scene is built from the first observed frame and is then consecutively transformed to the current observation based on the estimated \emph{model-to-frame} warp field at every time instance. This process generates a dense set of point trajectories, each associated with a point in the initial model. In order to keep the presentation clean, we opted to obtain the scene model from the first frame and keep it fixed in terms of its point set. Non-rigid reconstruction techniques for updating the model over time \cite{newcombe2015dynamicfusion,innmann2016volume} can be easily integrated to our pipeline if required.

To perform actor/background segmentation, we follow the semi-automatic approach described in Section \ref{sec:human_segmentation}. The obtained binary labeling is propagated to the whole temporal extent of the observed action via our estimated dense point trajectories, and enables us to easily detect human-environment \emph{contacts} as described in Section \ref{sec:contact_detection}.

Given the dense scene point trajectories, the actor/background labels, and the (hand) contact interaction locations and time intervals, our final goal is, for each detected interaction, to \emph{segment} the manipulated object and re-estimate its \emph{motion} for every time instance, assuming it is rigid (i.e. fully defined by a 6DOF pose). Our contact-guided motion segmentation approach for this task is described in Section \ref{sec:object_segmentation}.

In Table \ref{tab:method_i_o}, we summarize our proposed system's expected inputs, final outputs, and some useful generated intermediate results.

\begin{table}[!ht]
\centering
\caption{List of the inputs, intermediate results, and final outputs of our proposed system.}
\label{tab:method_i_o}
\scriptsize
\begin{tabular}{|c|c|c|}
\hline
\textbf{Input} & \textbf{Intermediate results} & \textbf{Final outputs}\\ \hline
\parbox[c]{0.17\textwidth}{\textbf{RGBD video} of manipulation} &
\parbox[c]{0.34\textwidth}{\begin{itemize}[nosep,noitemsep,wide=0pt]\item \textbf{Dense 3D point trajectories} for the whole sequence duration\item \textbf{Actor/background labels} for all model points at all times\end{itemize}} &
\parbox[c]{0.39\textwidth}{\vspace{0.4\baselineskip}\begin{itemize}[nosep,noitemsep,wide=0pt]\item \textbf{3D trajectories} of detected actor-environment \textbf{contact points}\item Manipulated \textbf{object segments} and their \textbf{6DOF poses} for every time point\end{itemize}\vspace{0.4\baselineskip}}
\\ \hline
\end{tabular}
\end{table}

\subsection{Non-rigid registration}\label{sec:non_rigid_icp}
As described in the previous subsection, whenever a new RGBD frame (point cloud) becomes available, our scene model is non-rigidly warped from its previous state (that corresponds to the previous frame) to the new (current) observation. Since parts of the scene model may be invisible in the current state (e.g., because of self-occlusion), we cannot directly apply a traditional scene flow algorithm, as that would only provide us with motion estimates for (some of) the currently visible points. Instead, we adopt a more general approach, by implementing a non-rigid Iterative Closest Point (ICP) algorithm, similar to \cite{amberg2007optimal, newcombe2015dynamicfusion,innmann2016volume}.

As is the case with rigid ICP \cite{besl1992method}, our algorithm iterates between a correspondence search step and a warp field optimization step for the given correspondences. Our correspondence search typically amounts to finding the nearest neighbors of each point in the current frame to the model point cloud in its previous state. Correspondences that exhibit large point distance, normal angle, or color difference are discarded. Nearest neighbor searches are done efficiently by parallel kd-tree queries.

In the following, we will focus on the warp field optimization step of our scheme. It has been found that modeling the warp field using locally affine \cite{amberg2007optimal} or locally rigid \cite{newcombe2015dynamicfusion} transformations provides better motion estimation results than adopting a simple translational local model, due to better regularization. In our implementation, for each point of the scene model in its previous state, we compute a full 6DOF rigid transformation that best aligns it to the current frame.

Let $X=\{x_i\}$ be the set of scene model points in the previous state that need to be registered to the point set $Y=\{y_i\}$ of the current frame, whose surface normals we denote by $Y^n=\{n_i\}$.
Let $S=\{s_i\}\subseteq\{1,\ldots,|X|\}$ and $D=\{d_i\}\subseteq \{1,\ldots,|Y|\}$ be the index sets of corresponding points in $X$ and $Y$ respectively, such that $(x_{s_i},y_{d_i})$ is a pair of corresponding points.
Let $T=\{T_i\}$ be the unknown warp field of rigid transformations, such that $T_i\in SE(3)$ and $|T|=|X|$, and $T_i(x_i)$ denote the application of $T_i$ to model point $x_i$.
Local transformations are parameterized by 3 Euler angles $(\alpha, \beta, \gamma)$ for their rotational part and 3 offsets $(t^x,t^y,t^z)$ for their translational part, and are represented as 6D vectors $T_i= \begin{bmatrix}\alpha_i & \beta_i & \gamma_i & t^x_i & t^y_i & t^z_i\end{bmatrix}^T$.

Our goal at this stage is to estimate a warp field $T$, of $6|X|$ unknown parameters, that maps model points in $S$ as closely as possible to frame models in $D$.
We formulate this property as the minimization of a weighted combination of sums of point-to-plane and point-to-point squared distances between corresponding pairs:
\begin{equation}
E_{\textrm{data}}(T) = \sum_{i=1}^{|S|}\left( n_{d_i}^T \left( T_{s_i}(x_{s_i}) - y_{d_i} \right)\right)^2 +
w_{\textrm{point}}\sum_{i=1}^{|S|} \left\lVert T_{s_i}(x_{s_i}) - y_{d_i}\right\rVert^2.
\label{eq:data_term}
\end{equation}
Pure point-to-plane metric optimization generally converges faster and to better solutions than pure point-to-point \cite{rusinkiewicz2001efficient} and is the standard trend in the state of the art for both rigid \cite{newcombe2011kinectfusion,whelan2015elasticfusion} and non-rigid \cite{newcombe2015dynamicfusion,innmann2016volume} registration. However, we have found that integrating a point-to-point term (second term in \eqref{eq:data_term}) with a small weight (e.g., with $w_{\textrm{point}}\approx 0.1$) to the registration cost improves motion estimation on surfaces that lack geometric texture.

The set of estimated correspondences is only expected to cover a subset of $X$ and $Y$, as not all model points are expected to be visible in the current frame, and the latter may suffer from missing data. Furthermore, even for model points with existing data terms (correspondences) in \eqref{eq:data_term}, analogously to the aperture problem in optical flow estimation, the estimation of point-wise transformation parameters locally is under-constrained. These reasons render the minimization of the cost function in \eqref{eq:data_term} ill-posed. To overcome this, we introduce a ``stiffness" regularization term that imposes an as-rigid-as-possible prior \cite{sorkine2007rigid} by directly penalizing differences between transformation parameters of neighboring model points in a way similar to \cite{amberg2007optimal}.
We fix a neighborhood graph on $X$, based on point locations, and use $\mathcal{N}(i)$ to denote the indices of the neighbors of point $x_i$ to formulate our stiffness prior term as:
\begin{equation}
E_{\textrm{stiff}}(T) = \sum_{i=1}^{|X|}\sum_{j\in \mathcal{N}(i)}w_{ij}\psi_\delta\left( T_i - T_j\right),
\label{eq:regularization}
\end{equation}
where $w_{ij} = \textrm{exp}\left(-\left\lVert x_i-x_j\right\rVert^2/(2\sigma_{\textrm{reg}}^2)\right)$, $\sigma_{\textrm{reg}}$ controls the radial extent of the regularization neighborhoods, `$-$' denotes regular matrix subtraction for the 6D vector representations of the local transformations, and $\psi_\delta$ denotes the sum of the Huber loss function values over the 6 residual components. Parameter $\delta$ controls the point at which the loss function behavior switches from quadratic ($L^2$-norm) to absolute-linear ($L^1$-norm).
Since $L^1$-norm regularization is known to better preserve solution discontinuities, we choose a small value of $\delta=10^{-4}$.

Our complete registration cost function is a weighted combination of costs \eqref{eq:data_term} and \eqref{eq:regularization}:
\begin{equation}
E(T) = E_{\textrm{data}}(T) + w_{\textrm{stiff}}E_{\textrm{stiff}}(T),
\label{eq:total_cost}
\end{equation}
where $w_{\textrm{stiff}}$ controls the overall regularization weight (set to $w_{\textrm{stiff}}=200$ in our experiments).
We minimize $E(T)$ in \eqref{eq:total_cost}, which is non-linear in the unknowns, by performing a small number of Gauss-Newton iterations. At every step, we linearize $E(T)$ around the current solution and obtain a solution increment $\hat{x}$ by solving the system of normal equations $J^TJ\hat{x}=J^Tr$, where $J$ is the Jacobian matrix of the residual terms in $E$ and $r$ is the vector of residual values.
We solve this sparse system iteratively, using the Conjugate Gradient algorithm with a diagonal preconditioner.


\begin{figure}[!ht]
    \centering
    \includegraphics[width=\textwidth]{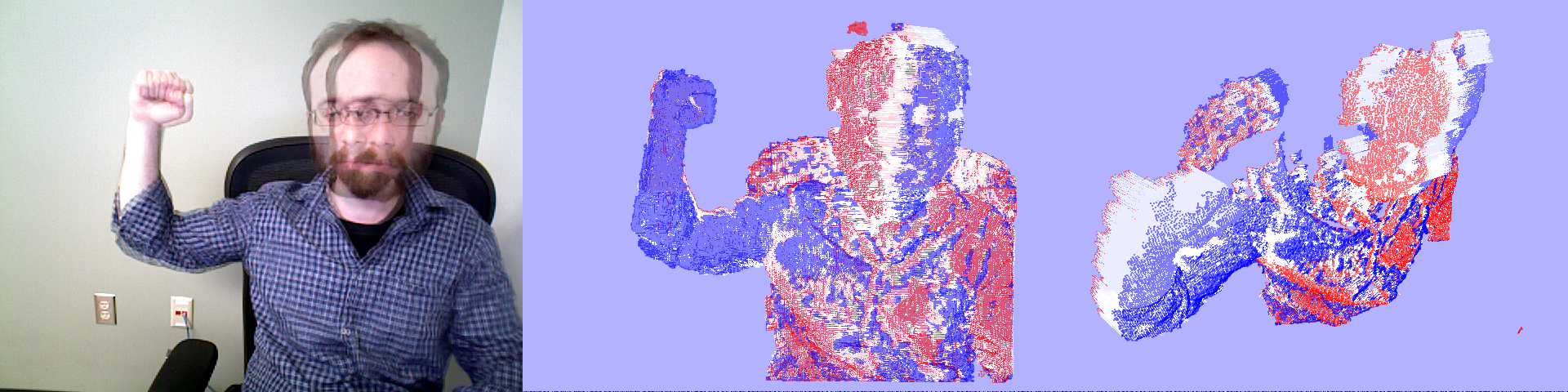}
    \includegraphics[width=\textwidth]{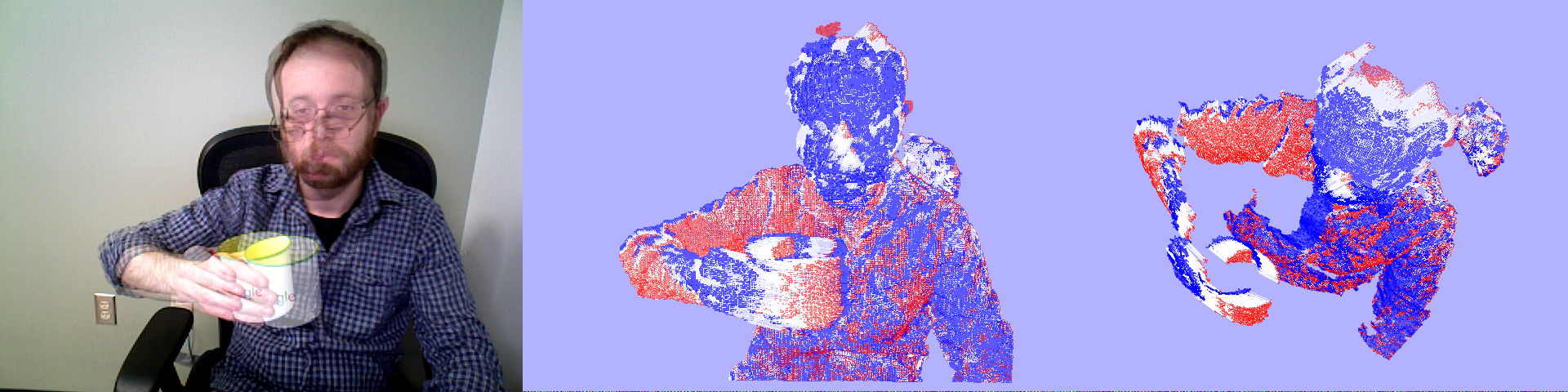}
    \caption{Non-rigid registration: displacement vectors are depicted as white lines, aligning the source (red) to the target (blue) geometry.}
    \label{fig:registration}
\end{figure}

In Fig. \ref{fig:registration}, we show two sample outputs of our algorithm in an RGBD frame pair non-rigid alignment scenario.
Our registration module accurately estimates deformations even for complex motions of significant magnitude.

\subsection{Human actor segmentation}\label{sec:human_segmentation}
We follow a semi-automatic approach to perform actor/background segmentation that relies on simple point cloud segmentation techniques.

We construct a proximity graph over the scene model points in the initial state, in which each node is a model point and two nodes are connected if and only if their Euclidean distance falls below a predefined threshold. 
Assuming that the actor is \emph{initially} not in contact with any other part of the scene (i.e. the minimum distance of an actor point to a background point is at least our predefined distance threshold) and the observed actor points are not too severely disconnected in the initial state, the actor points will be exactly defined by one connected component of this proximity graph.
The selection of the correct (actor) component can be automated by filtering all the extracted components based on context-specific criteria (e.g., rough size, shape, location, etc.) or by picking the component whose image projection exhibits maximum overlap with the output of a 2D human detector \cite{cao2017realtime,dalal2005histograms}.
Equivalently, we may begin by selecting a seed point known to belong to the actor and then perform region growing on the model point cloud until our distance threshold is no longer satisfied. Again, the selection of the seed point can be automated by resorting to standard 2D means (e.g., by picking the point with the strongest skin color response \cite{jones2002statistical,vezhnevets2003survey} within a 2D human detector output \cite{cao2017realtime,dalal2005histograms}).

We believe that the assumptions imposed by our Euclidean clustering based approach for the actor segmentation task are not too restricting, as the main setting we focus on (representing human demonstrations for robot learning) is reasonably controlled in the first place.

We note that, since we opted to keep the scene model point set fixed and track it throughout the observed action, the obtained segmentation automatically becomes available at all time points.

\subsection{Contact detection}\label{sec:contact_detection}
The outputs of the above two processes are a dense set of \emph{point trajectories} and their respective actor/background \emph{labels}. Given this information, it is straightforward to reason about \emph{contact}, simply by examining whether the minimum distance between parts of the two clusters is small enough at any given time. In other words, we can easily infer both \emph{when} the actor comes into/goes out of contact with part of the environment and \emph{where} this interaction is taking place.

Some of the contact interactions detected using this criterion may, of course, be semantically irrelevant to the performed action. Since semantic reasoning is not part of our core framework, these cases have to be handled by a higher lever module. However, under reasonably controlled scenarios, we argue that it is sufficient to simply assume that the detected contacts are established by the actor \emph{hands}, with the goal of manipulating an \emph{object} in their environment.

\subsection{Manipulated object motion/segmentation}\label{sec:object_segmentation}
Knowing the dense scene point trajectories, labeled as either actor or background, as well as the contact locations and intervals, our next goal is to infer what part of the environment is being manipulated, or, in other words, which object was moved. We assume that every contact interaction involves the movement of a \emph{single} object, and that the latter undergoes \emph{rigid} motion. In the following, we only focus on the \emph{background} part of the scene around the contact point area, ignoring the human point trajectories. We propose the following two-step approach.

First, we bootstrap our segmentation task by finding a coarse/partial mask of the moving object, using standard unsupervised clustering techniques. Specifically, we cluster the point trajectories that are labeled as background and lie within a fixed radius of the detected contact point at the beginning of the interaction into two groups. We adopt a spectral clustering approach, using the `random walk' graph Laplacian \cite{von2007tutorial} and a standard $k$-means last step. Our pairwise trajectory similarities are given by $s_{ij}=\mathrm{exp}\left(-(d_{max}-d_{min})^2/(2\sigma^2)\right)$, where $d_{min}$ and $d_{max}$ are the minimum and maximum Euclidean point distance of trajectories $i$ and $j$ over the duration of the interaction, respectively. This similarity metric enforces similar trajectories to exhibit relatively constant point-wise distances, i.e. promotes clusters that undergo rigid motion.
From the two output clusters, one is expected to cover (part of) the object being manipulated. Operating under the assumption that only interaction can cause motion in the scene, we pick the cluster that exhibits the largest average motion over the duration of contact as our object segment candidate.

In the above, we restricted our focus within a region of the contact point, in order to \emph{1)} avoid that our binary classification is influenced by other captured motions in the scene that are not related to the current interaction, and \emph{2)} make the classification itself more computationally tractable. As long as these requirements are met, the choice of radius is not important.

Subsequently, we obtain a refined, more accurate segment of the moving object by requiring that the latter undergoes a rigid motion that is at every time point consistent with that of the previously found motion cluster.
Let $B^t$ denote the background (non-actor) part the scene model point cloud at time $t$, for $t=0,\ldots,T$, and $\hat{M}^t\subseteq B^t$ be the initial motion cluster state at the same time instance.
For all $t=1,\ldots,T$, we robustly estimate the rigid motion between point sets $\hat{M}^0$ and $\hat{M}^t$ (i.e. relative to the first frame), using the closed form solution of \cite{umeyama1991least} under a RANSAC scheme, and then find the set of points in \emph{all} of $B^t$ that are consistent with this motion model between $B^0$ and $B^t$.
If we denote this set of motion inliers by $I^t$ (which is a set of indices of points in $B^t$), we obtain our final object segment for this interaction as the intersection of inlier indices for all time instances $t=1,\ldots,T$:
\begin{equation}
I\equiv\bigcap_{t=1}^{T}I^t
\end{equation}
The subset of the background points indexed by $I$, as well as the per-frame RANSAC motion (pose) estimates of this last step, are the final outputs of our pipeline for the given interaction.

\newpage

\section{Experiments}\label{sec:experiments}

\subsection{Qualitative evaluation}\label{sec:qualitative_evaluation}
We provide a qualitative evaluation of our method for video inputs recorded in different settings, covering three different scenarios: \emph{1)} a tabletop object manipulation that involves flipping a pitcher, \emph{2)} opening a drawer, and \emph{3)} opening a room door.
All videos were captured from a static viewpoint, using a standard RGBD sensor.

\begin{figure}[!htp]
	\centering
    \includegraphics[width=0.85\textwidth]{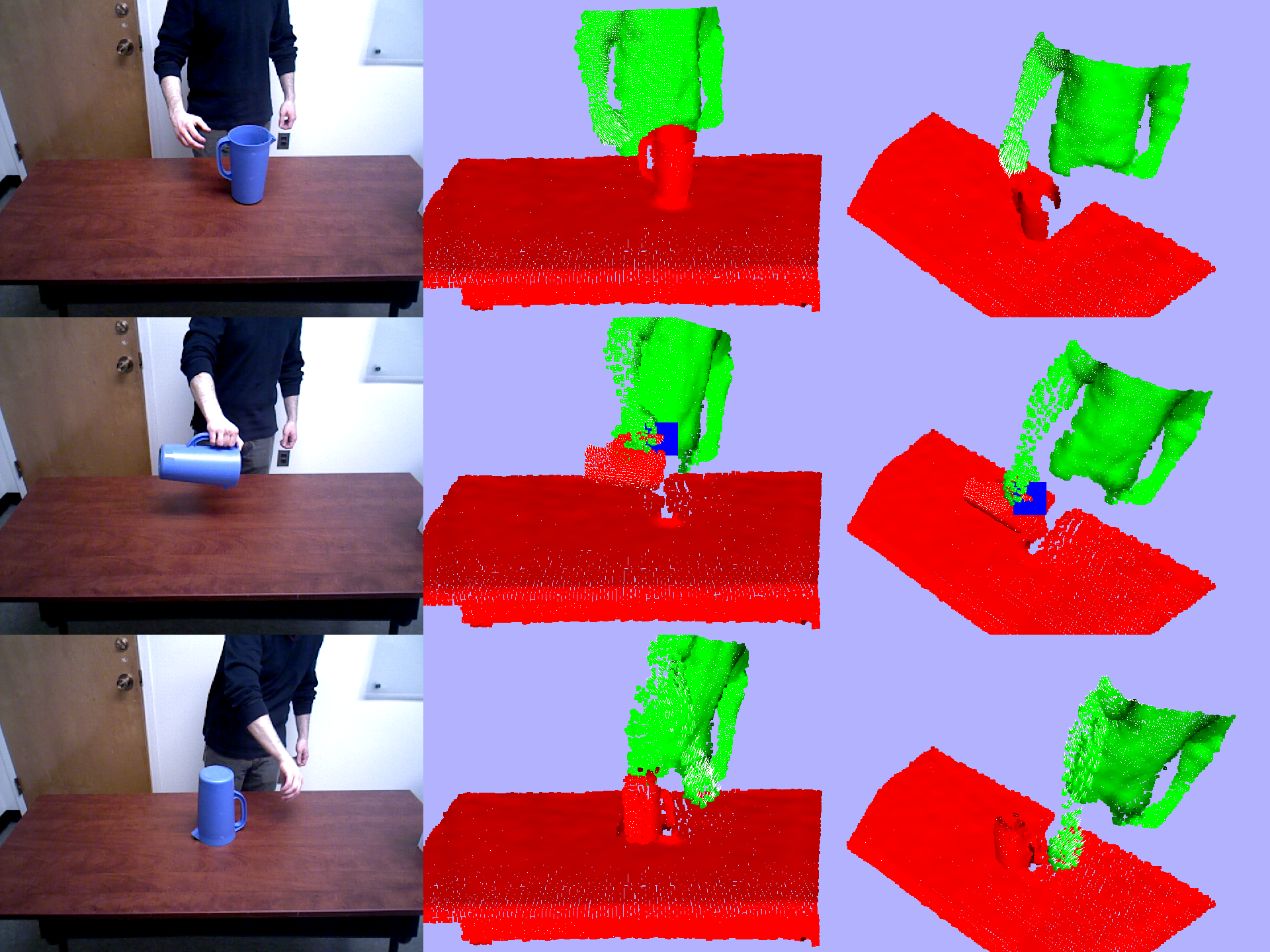}
    \caption{Flipping a pitcher: scene tracking, labeling, and contact detection.}
    \label{fig:pitcher_tracking}
\end{figure}

\begin{figure}[!htp]
	\centering
    \includegraphics[width=0.85\textwidth]{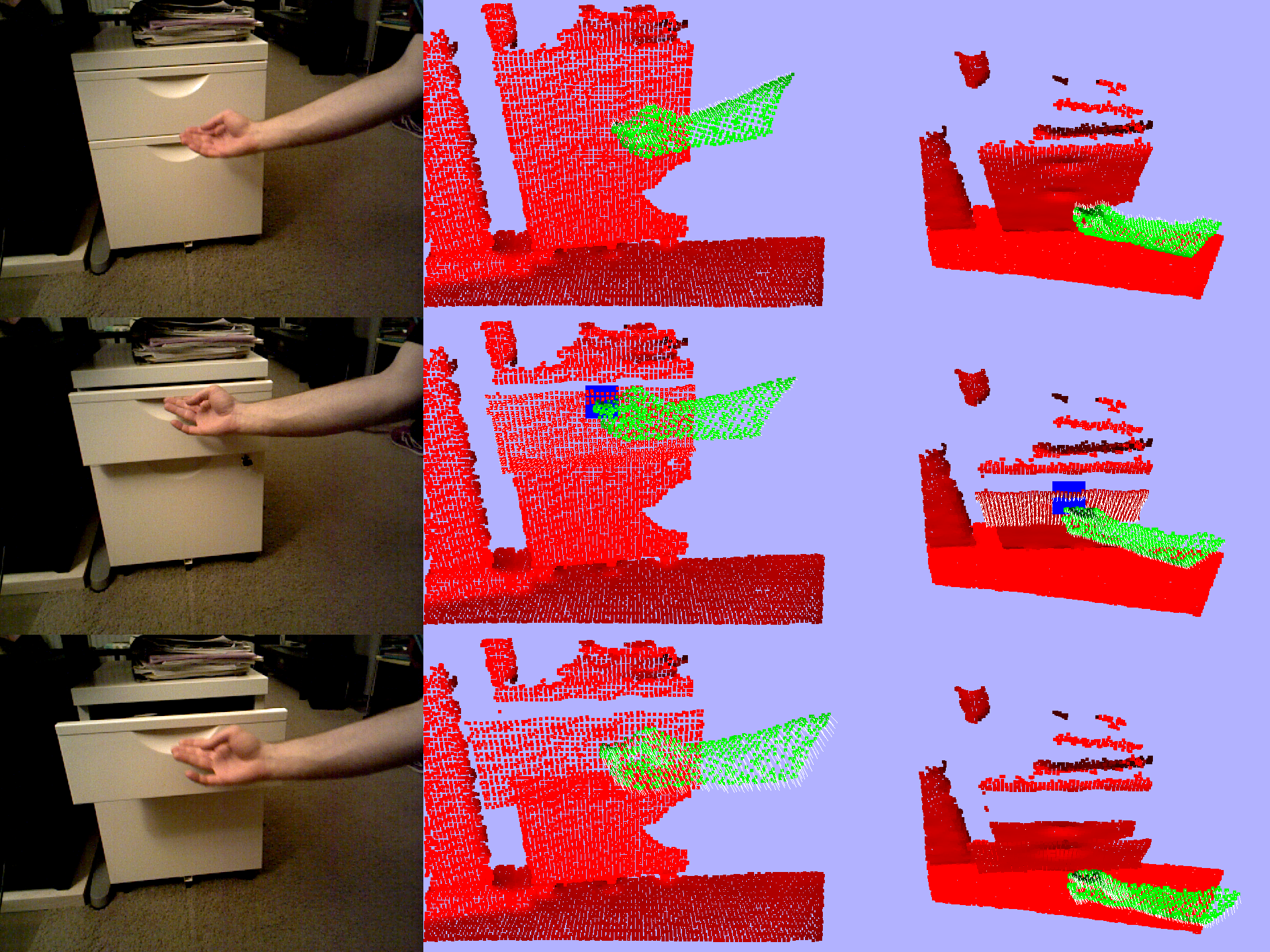}
    \caption{Opening a drawer: scene tracking, labeling, and contact detection.}
    \label{fig:top_drawer_tracking}
\end{figure}

\begin{figure}[!htp]
	\centering
    \includegraphics[width=0.85\textwidth]{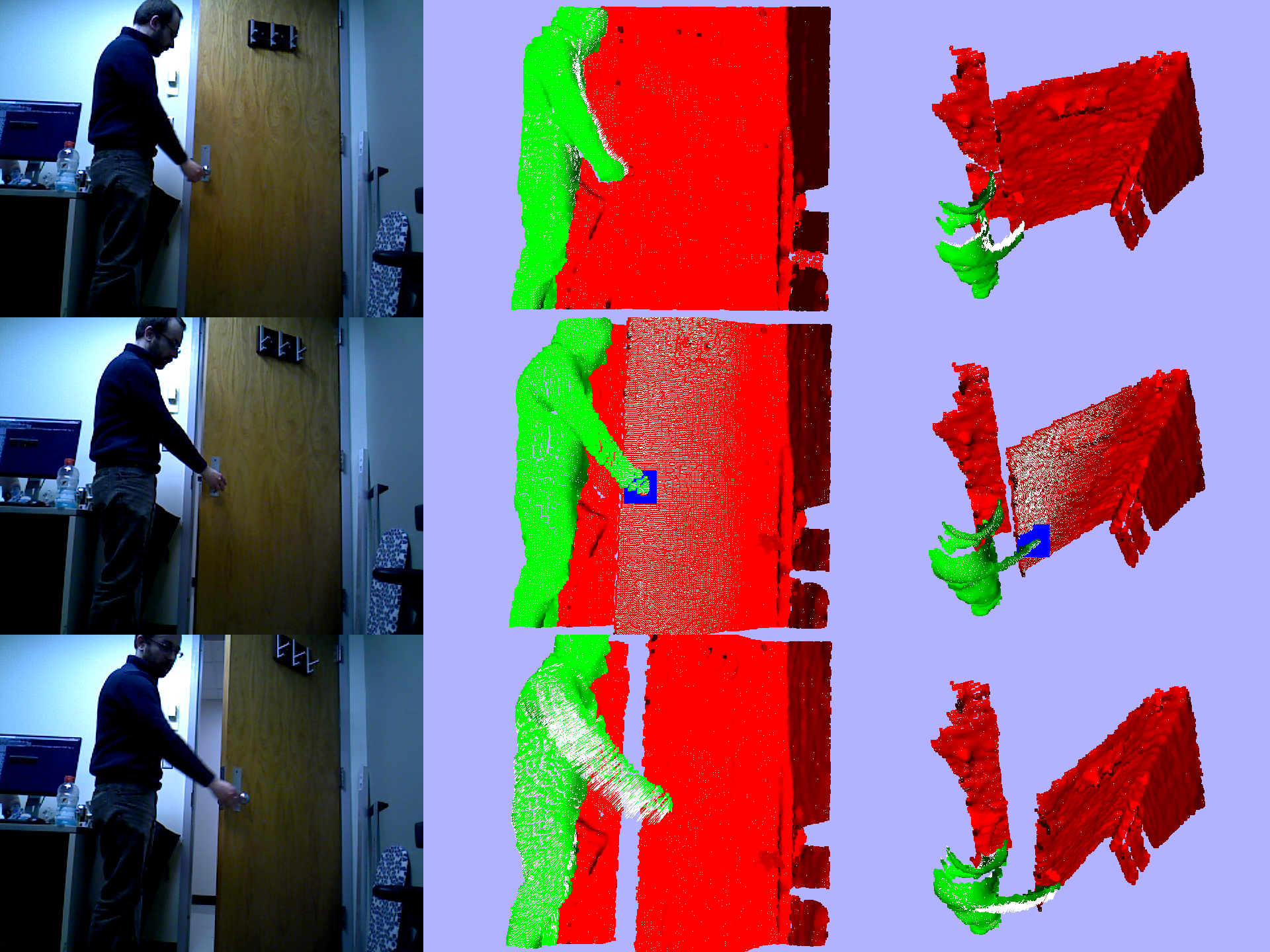}
    \caption{Opening a door: scene tracking, labeling, and contact detection.}
    \label{fig:office_door_tracking}
\end{figure}

For each scenario, we depict (in Fig. \ref{fig:pitcher_tracking}, \ref{fig:top_drawer_tracking}, and \ref{fig:office_door_tracking}, respectively) the scene model point cloud state at three time snapshots: one right before, one during, and one right after the manipulation.
For each time point, we show the corresponding color image and render the tracked point cloud from two viewpoints.
The actor segment is colored green, the background is red, and the detected contact area is marked by blue.
We also render the point-wise displacements induced by the estimated warp field (from the currently visible state to its next) as white lines (mostly visible in areas that exhibit large motion).
The outputs displayed in these figures are in direct correspondence with the processes described in Sections \ref{sec:non_rigid_icp}, \ref{sec:human_segmentation}, and \ref{sec:contact_detection}.

\strut
\newpage

\begin{figure}[!hp]
    \centering
    \begin{subfigure}[b]{\textwidth}
        \includegraphics[width=\textwidth]{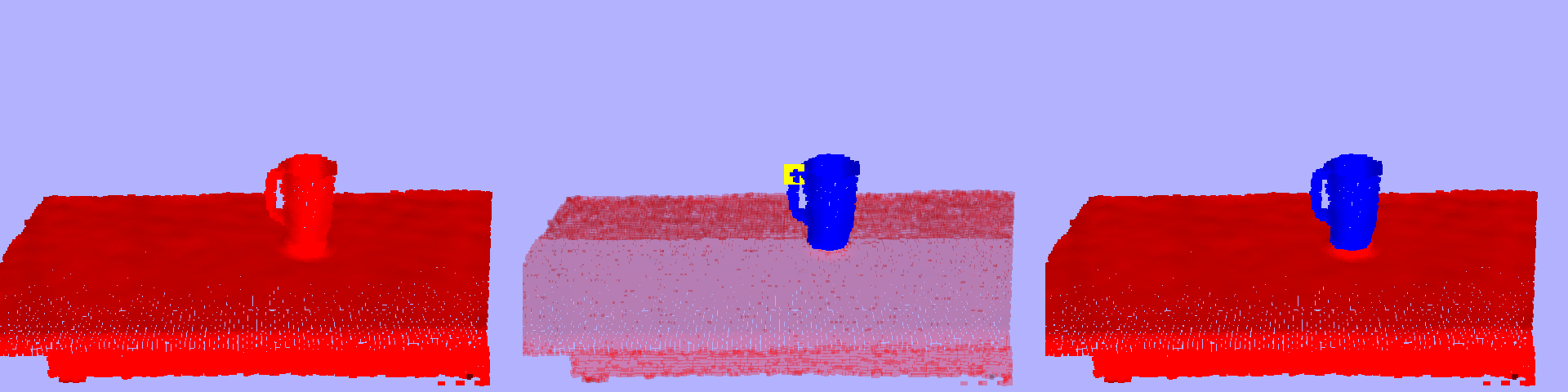}
        \caption{Flipping a pitcher.}
        \label{fig:pitcher_segmentation}
    \end{subfigure}
    \begin{subfigure}[b]{\textwidth}
        \includegraphics[width=\textwidth]{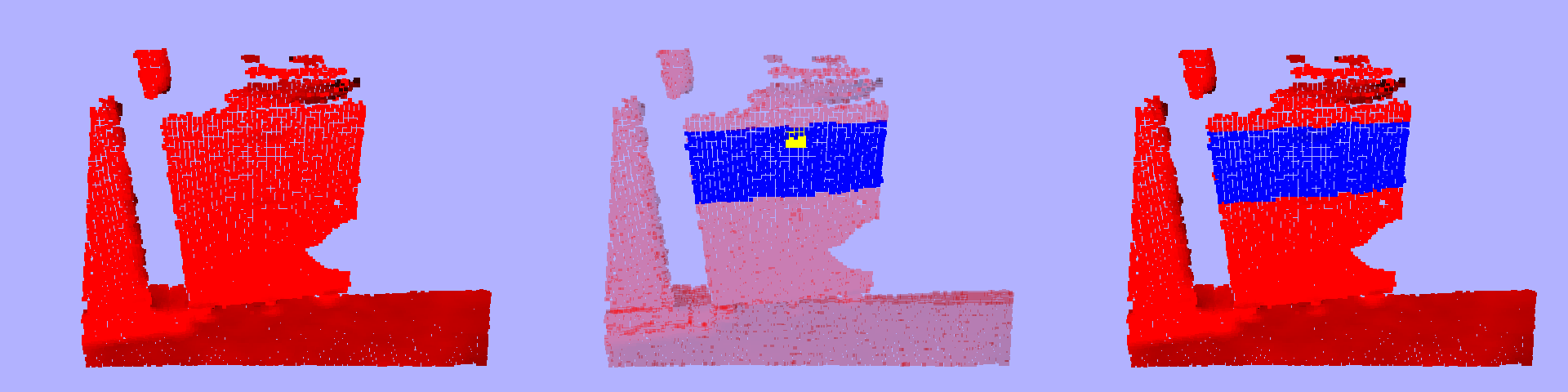}
        \caption{Opening a drawer.}
        \label{fig:top_drawer_segmentation}
    \end{subfigure}
    \begin{subfigure}[b]{\textwidth}
        \includegraphics[width=\textwidth]{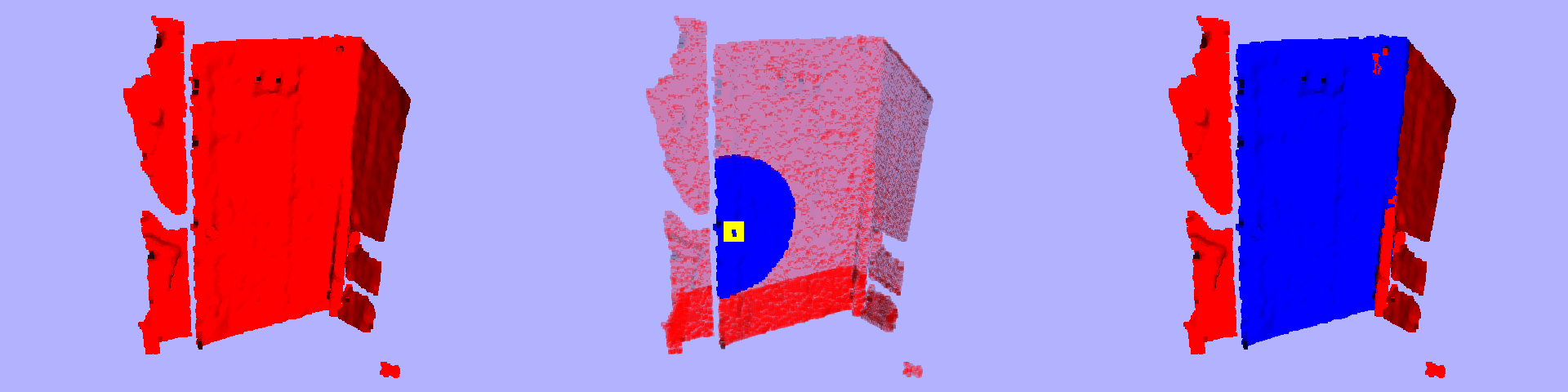}
        \caption{Opening a door.}
        \label{fig:office_door_segmentation}
    \end{subfigure}
    \caption{Motion segmentation of the manipulated object. First column: scene background points (the actor is removed). Second column: initial motion segment (blue) obtained by spectral clustering of point trajectories around contact area (yellow). Third column: Final motion segment.}
    \label{fig:motion_segmentation}
\end{figure}

Next, we demonstrate our attention-driven motion segmentation and 6DOF pose estimation of the manipulated object. In Fig. \ref{fig:motion_segmentation}, we render the background part of the scene model in its initial state with the actor removed and show the two steps of our segmentation method described in Section \ref{sec:object_segmentation}.
In the middle column, the blue segment corresponds to the initial motion segment, obtained by clustering trajectories in the vicinity of the contact point, which was propagated back to the initial model state and is highlighted in yellow.
In the left column, we show the refined, final motion segment.
We note that, because of our choice of the radius around the contact point in which we focus out attention in the first step, the initial segment in the first two cases is the same as the final one.

\newpage

\begin{figure}[!hp]
    \centering
    \begin{subfigure}[b]{\textwidth}
        \includegraphics[width=\textwidth]{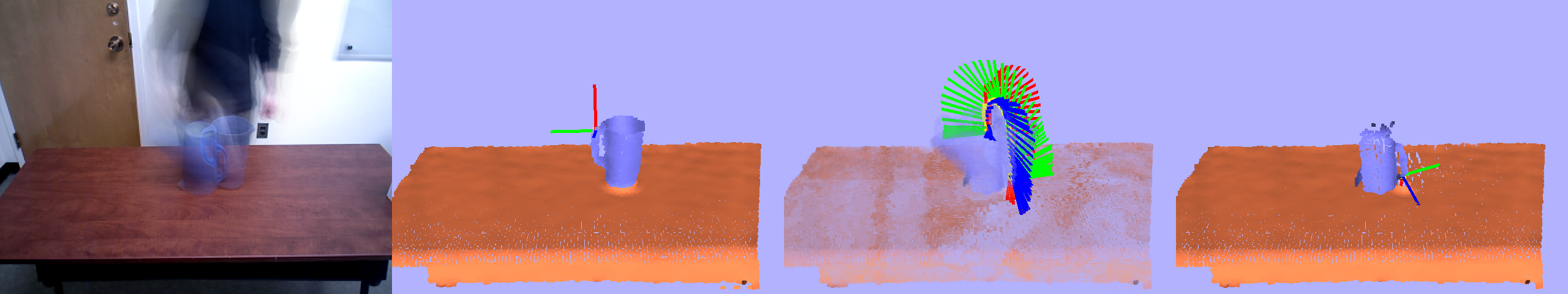}
        \caption{Flipping a pitcher.}
        \label{fig:pitcher_trajectory}
    \end{subfigure}
    \begin{subfigure}[b]{\textwidth}
        \includegraphics[width=\textwidth]{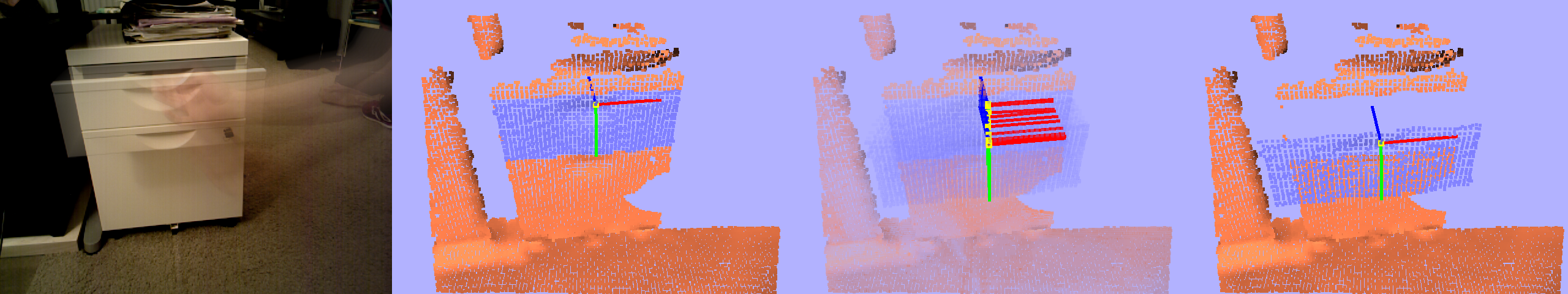}
        \caption{Opening a drawer.}
        \label{fig:top_drawer_trajectory}
    \end{subfigure}
    \begin{subfigure}[b]{\textwidth}
        \includegraphics[width=\textwidth]{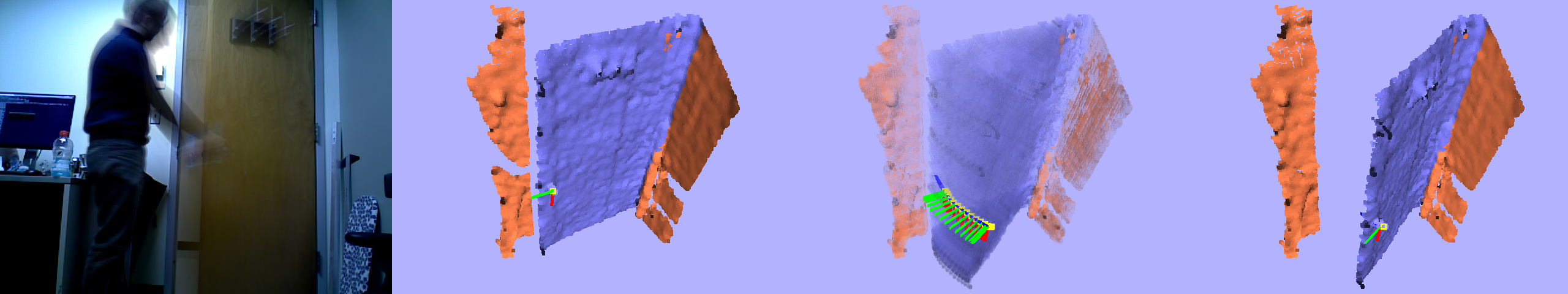}
        \caption{Opening a door.}
        \label{fig:office_door_trajectory}
    \end{subfigure}
    \caption{Estimated rigid motion of the manipulated object. A coordinate frame is attached to the object segment (blue) at the contact point location (yellow). First column: temporal accumulation of color frames for the whole action duration. Second column: object state before manipulation. Third column: object trajectory as a series of 6DOF poses. Fourth column: object state after manipulation.}
    \label{fig:motion_trajectory}
\end{figure}

In Fig. \ref{fig:motion_trajectory}, we show the estimated rigid motion (6DOF pose) of the segmented object.
To more clearly visualize the evolution of object pose over time, we attach a local coordinate frame to the object, at the location of the contact point, whose axes were chosen as the principal components of the extracted object point cloud segment.

The above illustrations provide a qualitative demonstration of the successful application of our proposed pipeline to three different manipulation videos. In all cases, contacts were detected correctly and the manipulated object was accurately segmented and tracked. A more thorough, quantitative evaluation of our contact and segmentation outputs on an extended set of videos is in our plans for the immediate future.

\subsection{Implementation}\label{sec:implementation}
Our pipeline is implemented using the \texttt{cilantro} \cite{cilantro} library, which provides a self-contained set of facilities for all of the computational steps involved.

\newpage

\section{Application: Replication from Observation by a robot}\label{sec:applications}
\begin{figure}[!ht]
	\centering
	\includegraphics[width=0.4\textwidth]{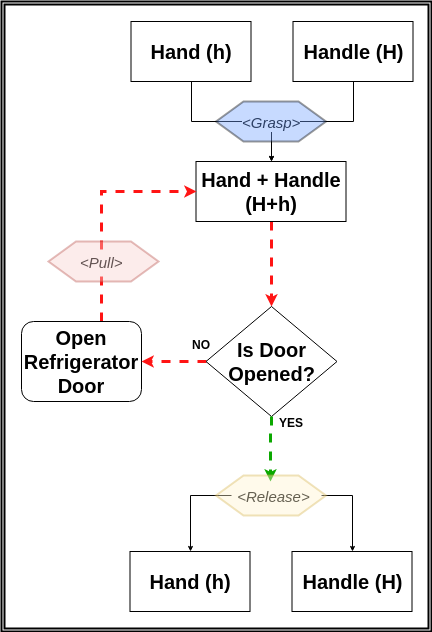}
    \caption{High-Level representation of opening a refrigerator door.}
    \label{fig:contact_graph}
\end{figure}
For any human-environment task to be successful, there is a well-defined process involved, demarcated into phases depending on human-environment contact and consequent motion. This allows us to generate a graph representation for actions, such as that shown in Fig.~\ref{fig:contact_graph}, for the task of opening a refrigerator. Given this general representation of tasks, we demonstrate how our algorithm allows grounding of the \textit{grasp} and \textit{release} parts, based on contact detection, and also of the feedback loop for opening the door, based on motion analysis of segmented objects. Such a representation, featuring a tight coupling of planning and perception, is crucial for robots to observe and replicate human actions.

\begin{figure}[!ht]
	\centering
	\includegraphics[width=0.57\textwidth]{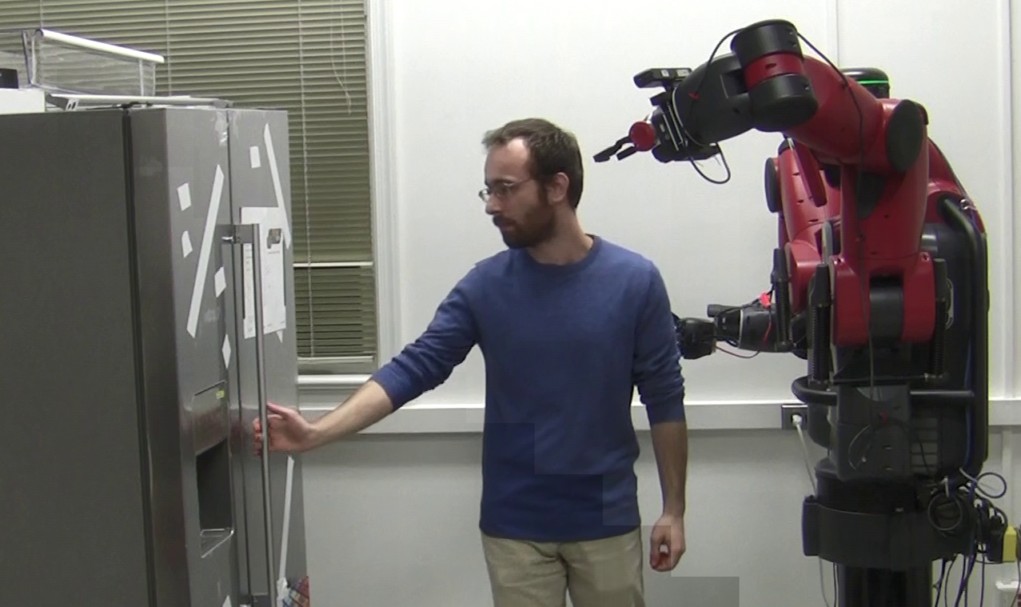}
    \caption{Robot observing a human opening a door.}
    \label{fig:robot_observing_human}
\end{figure}

We now present a comprehensive application of our method to a real-world task, where a robot observes a human operator opening a refrigerator door and learns the process for replication. This can be seen in Fig.~\ref{fig:robot_observing_human}, where a RGBD sensor mounted to the robot's manipulator is used for observation.
This process involves the segmentation of the human and the environment from the observed video input, analyzing the contact between the human agent and the environment (the refrigerator handle in this case), and finally performing 3D motion tracking and segmentation on the action of opening the door, using our methods elucidated in Section \ref{sec:approach}.
These analyses, and the corresponding outputs, are then converted into an intermediate graph-like representation, which encodes both semantic labeling of regions of interest, such as doors and handles in our case, as well as motion trajectories computed from observing the human agent. The combination of these allow the robot to understand and generalize the action to be performed even in changing scenarios.

\begin{figure}[!ht]
	\centering
	\includegraphics[width=0.7\textwidth]{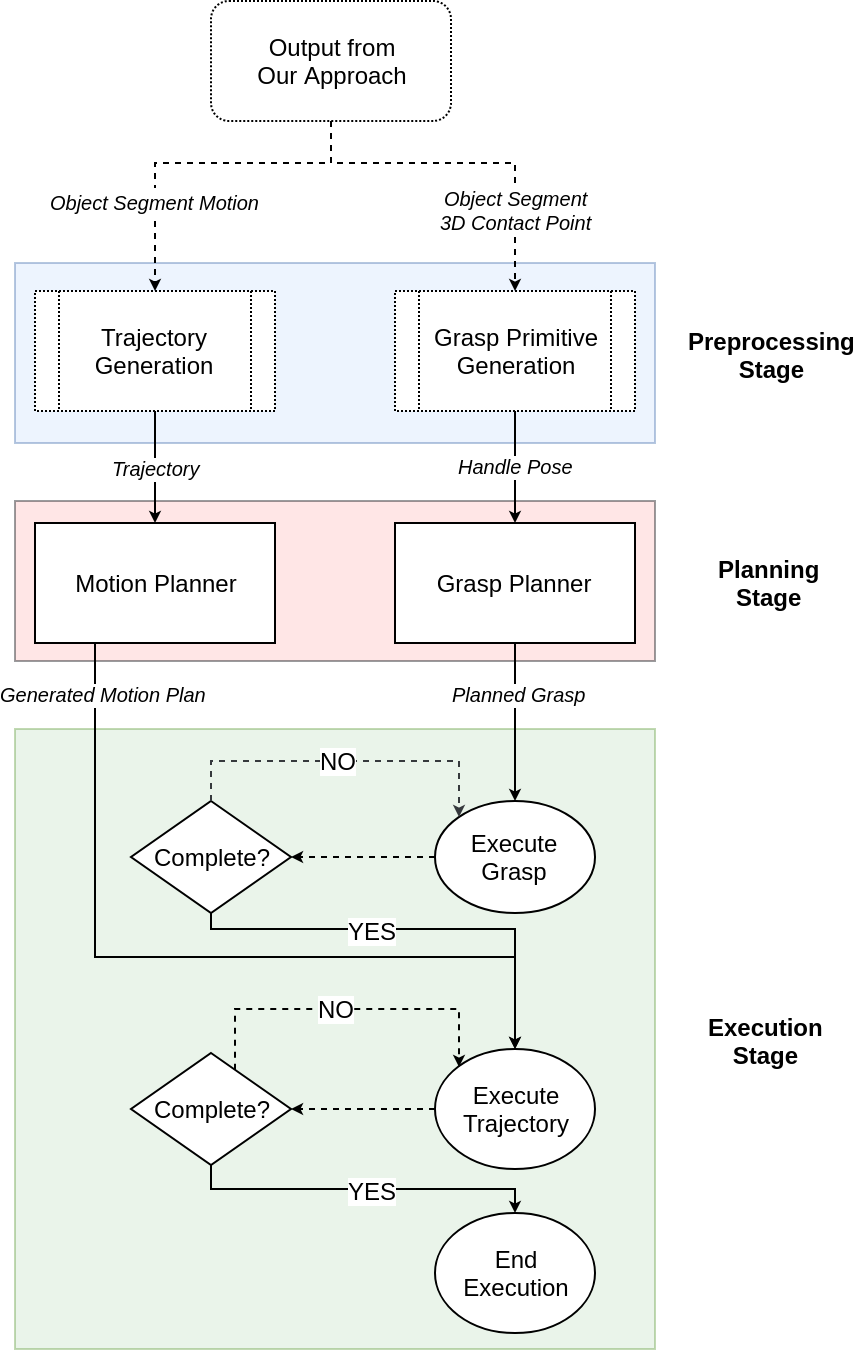}
    \caption{State transition diagram of our process.}
    \label{fig:transition_diagram}
\end{figure}

We present a detailed explanation of each step involved in the process of a robot's replication of an action by observing a human.
This entire process is visually described in Fig.~\ref{fig:transition_diagram}, which separates our application into three phases, namely \textit{preprocessing}, \textit{planning} and \textit{execution}.

\subsection{Preprocessing Stage}

\begin{figure}[!ht]
	\centering
	\includegraphics[width=1.0\textwidth]{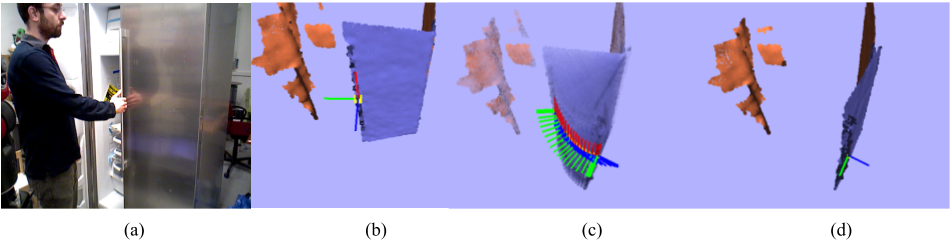}
    \caption{Input to the \textit{preprocessing} stage from our algorithm.}
    \label{fig:preprocessing_input}
\end{figure}

\begin{figure*}[!ht]
	\begin{subfigure}[t]{0.5\textwidth}
      \centering
      \includegraphics[width=0.6\columnwidth]{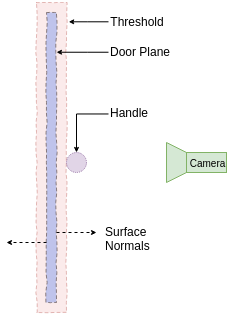}
      \caption{Diagram depicting refrigerator handle detection.}
      \label{fig:handle_detector}
	\end{subfigure}
    ~
    \begin{subfigure}[t]{0.5\textwidth}
      \centering
      \includegraphics[width=0.6\columnwidth]{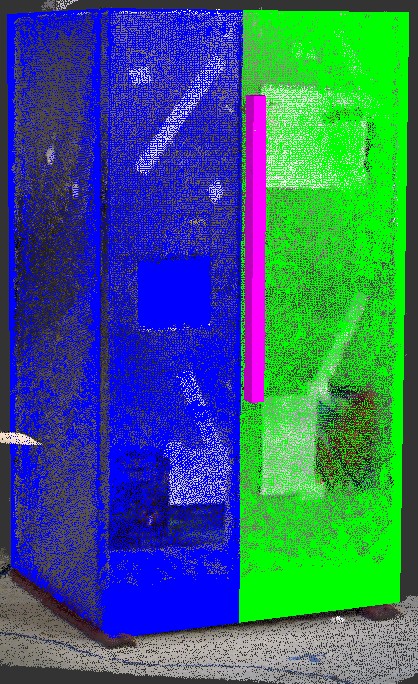}
      \caption{Point cloud of refrigerator with detected handle and door.}
      \label{fig:fridge_handle}
	\end{subfigure}
    \caption{Handle detection.}
\end{figure*}

The preprocessing stage is responsible for taking the contact point, object segments and their motion trajectories, as described in Fig.~\ref{fig:overview}, and converting them into robot-specific trajectories for planning and execution. A visualization of this input can be seen in Fig.~\ref{fig:preprocessing_input}, where \textbf{(a)} depicts the RGB frame of the human performing the action. Subfigure \textbf{(b)} shows the contact point, highlighted in yellow, along with an initial object frame. Subfigure \textbf{(c)} demonstrates a dynamic view of the motion trajectory and segmentation of the door, along with the tracked contact point axes across time. Subfigure \textbf{(d)} shows the final pose of the door, after opening has finished.

In this stage, we exploit domain knowledge to semantically ground contact points and object segments, in order to assist affordance analysis and common-sense reasoning for robot manipulation, since that provides us with task-dependent priors. For instance, since we know that our task involves opening a refrigerator door, we can make prior assumptions that the contact point between the human agent and the environment will happen at the handle and any consequent motion will be of the door and handle only.

\subsubsection{Door Handle Detection}\label{subsub:handle}
These priors allow us to robustly fit a plane to the points of the door (extracted object) using standard least squares fitting under RANSAC and obtain a set of points for the door handle (plane outliers). We then fit a cylinder to these points, in order to generate a grasp primitive with a 6 DOF pose, for robot grasp planning.
The estimated trajectories of the object segment, as mentioned in Table~\ref{tab:method_i_o}, are not directly utilized by the robot execution system, but must instead be converted to a robot-specific representation before replication can take place. Our algorithm outputs a series of 6DOF poses $P_i$ for every time point $t_i \in T$. These are then converted to a series of robot-usable poses for the planning phase.

\subsection{Planning Stage}
\begin{figure}[!ht]
	\centering
	\includegraphics[width=0.7\textwidth]{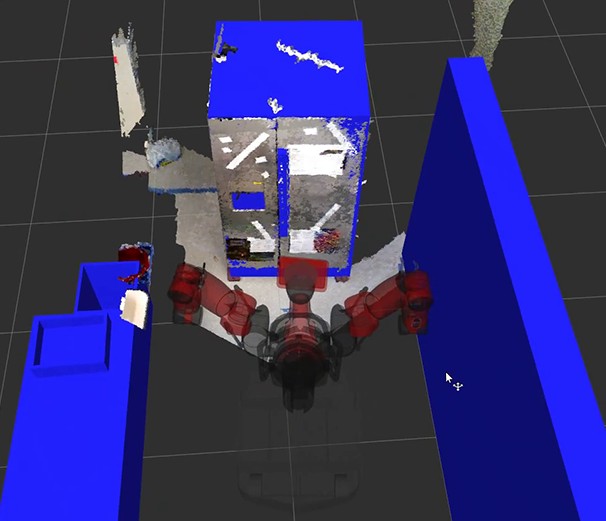}
    \caption{Visualization of planning stage}
    \label{fig:rviz}
\end{figure}
The outputs from the preprocessing stage, namely the robot-specific 6DOF poses of the handle and the cylinder of specified radius and height depicting the handle are passed in to the \textit{planning} stage of our pipeline, for both grasp planning and trajectory planning. The Robot Visualizer (rviz) \cite{rviz} package in ROS allows for simulation and visualization of the robot during planning and execution, via real-time feedback from the robot's state estimator. It also has point cloud visualization capabilities, which can be overlaid over primitive shapes. We use this tool for the planning stage, with the Baxter robot and our detected refrigerator.
\subsubsection{Grasp Planning}
Given a primitive shape, such as a block or cylinder, we are able to use the MoveIt! Simple Grasps \cite{moveitgrasp2016} package to generate grasp candidates for a parallel gripper (such as one mounted on the Baxter robot). The package integrates with the ``MoveIt!'' library's pick and place pipeline to simulate and generate multiple potential grasp candidates, i.e. approach poses. There is also a grasp filtering stage, which uses task and configuration specific constraints to remove kinematically infeasible grasps, by performing feasibility tests via inverse kinematics solvers. At the end of the grasp planning pipeline, we have a set of candidate grasps, sorted by a grasp quality metric, of which one is chosen for execution in the next stage.
\subsubsection{Trajectory Planning}
The ordered set of the poses over time obtained from the \textit{preprocessing} stage is then used to generate a Cartesian path, using the Robot Operating System's ``MoveIt!'' \cite{chitta2012moveit} motion planning library. This abstraction allows us to input a set of poses through which the end-effector must pass, along with parameters for path validity and obstacle avoidance. ``MoveIt!'' then uses inverse kinematics solutions for the specified manipulator configuration combined with sampling-based planning algorithms, such as Rapidly-Exploring Random Trees \cite{lavalle98rrt}, to generate a trajectory for the robot to execute.

\subsection{Execution Stage}
\begin{figure}[!ht]
	\centering
	\includegraphics[width=\textwidth]{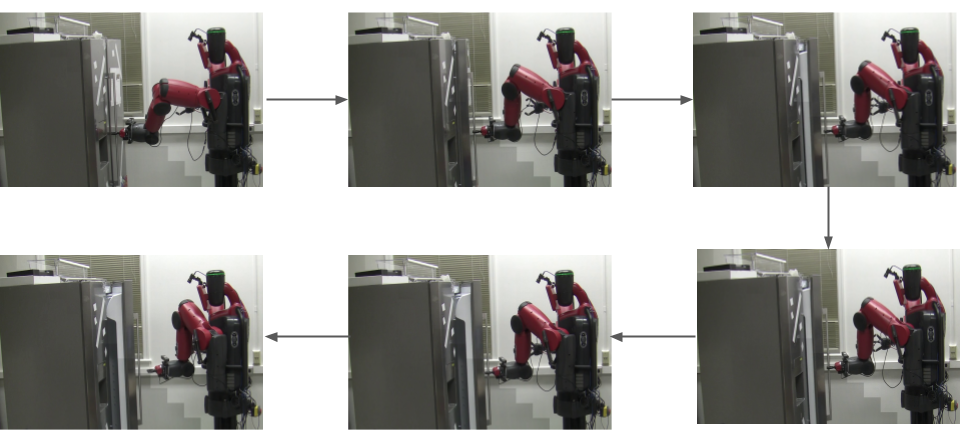}
    \caption{Robot replicating human by opening refrigerator}
    \label{fig:robot_opening_refrigerator}
\end{figure}
The \textit{execution} stage takes as input the grasp and trajectory plans generated in the \textit{planning} stage and executes the plan on the robot. First, the generated grasp candidate is used to move the end-effector to a pre-grasp pose and the parallel gripper is aligned to the cylindrical shape of the handle. The grasp is executed based on a feedback control loop, with the termination condition decided by collision avoidance and force feedback. Upon successful grasp of the handle, our pipeline transitions into the trajectory execution stage, which attempts to follow the generated plan based on feedback from the robot's state estimation system. Once the trajectory has been successfully executed, the human motion replication pipeline is complete.
This execution process is demonstrated by the robot in Fig.~\ref{fig:robot_opening_refrigerator}, beginning with the robot grasping the handle in the top-leftmost figure and ending with the robot releasing the handle in the bottom-leftmost figure, with intermediate frames showing the robot imitating the motion trajectory of the human.

In future work, we plan to implement a dynamic motion primitives \cite{schaal_dmp_2016} (DMP) based approach, which will allow more accurate and robust tracking of trajectories by the robot.

\section{Conclusions}\label{sec:conclusions}
In this paper, we have introduced an active, bottom-up method for the extraction of two fundamental features of an observed manipulation, namely the contact points and motion trajectories of segmented objects.
We have qualitatively demonstrated the success of our approach on a set of video inputs and described in detail its fundamental role in a robot imitation scenario.
Owing to its general applicability and the manipulation-defining nature of its output features, our method can effectively bridge the gap between observation and the development of action representations and plans.

There are many possible directions for future work.
At a lower level, we plan to integrate \emph{dynamic reconstruction} into our pipeline to obtain a more complete model for the manipulated object; at this moment, this can be achieved by introducing a step of static scene reconstruction before the manipulation happens, after which we run our algorithm.
We also plan to extend our method so that it also can handle \emph{articulated} manipulated objects, as well as objects that are \emph{indirectly} manipulated (e.g., via the use of tools).

On the planning end, one of our future goals is to release a software component for the fully automated replication of \emph{door opening} tasks (Section \ref{sec:applications}), given only a single demonstration. This module will be hardware agnostic up until the final execution stage of the pipeline, such that the generated plan to be imitated can be handled by any robot agent, given the specific manipulator and end-effector configurations.

\section*{Acknowledgments}
The support of ONR under grant award N00014-17-1-2622 and the support of the National Science Foundation under grants SMA 1540916 and CNS 1544787 are greatly acknowledged.

\bibliography{references}

\end{document}